\documentclass[10pt,twocolumn,letterpaper]{article}

\usepackage{graphicx}
\usepackage{algorithm}
\usepackage{listings}
\usepackage{multirow}
\usepackage{cvpr}              % To produce the CAMERA-READY version
\usepackage[dvipsnames]{xcolor}

\definecolor{cvprblue}{rgb}{0.21,0.49,0.74}
\usepackage{hyperref}
\hypersetup{pagebackref,breaklinks,colorlinks,citecolor=cvprblue}
\def\paperID{14637} % *** Enter the Paper ID here
\def\confName{CVPR}
\def\confYear{2024}

\title{Data-efficient Event Camera Pre-training via Disentangled Masked Modeling}

\author{Zhenpeng Huang$^1$, Chao Li$^2$, Hao Chen$^{1*}$, Yongjian Deng$^3$, Yifeng Geng$^2$, Limin Wang$^4$\\
$^1$Southeast University,$^2$Alibaba Group, $^3$Beijing University of Technology,$^4$Nanjing University\\
\\
\and
}

\begin{document}
\maketitle
\begin{abstract}
In this paper, we present a new data-efficient voxel-based self-supervised learning method for event cameras. Our pre-training overcomes the limitations of previous methods, which either sacrifice temporal information by converting event sequences into 2D images for utilizing pre-trained image models or directly employ paired image data for knowledge distillation to enhance the learning of event streams. In order to make our pre-training data-efficient, we first design a semantic-uniform masking method to address the learning imbalance caused by the varying reconstruction difficulties of different regions in non-uniform data when using random masking.
% To resolve the conflict between the scarcity of event data and the need for extensive unlabeled data in pre-training, 
In addition, we ease the traditional hybrid masked modeling process by explicitly decomposing it into two branches, namely local spatio-temporal reconstruction and global semantic reconstruction to encourage the encoder to capture local correlations and global semantics, respectively.
% Specifically, we consider the sparsity and uneven distribution of event data and identify that the conventional globally random masking strategy can cause imbalanced reconstruction and learning across different regions of event data. To rectify this issue, we propose a locally uniform masking method to ensure balanced learning for each region. 
% Moreover, we incorporate masking reconstruction in high-level features to encourage the encoder to capture global semantics. 
This decomposition allows our self-supervised learning method to converge faster with minimal pre-training data. Compared to previous approaches, our self-supervised learning method does not rely on paired RGB images, yet enables simultaneous exploration of spatial and temporal cues in multiple scales. It exhibits excellent generalization performance and demonstrates significant improvements across various tasks with fewer parameters and lower computational costs.
% , including object recognition, object detection, semantic segmentation, and action recognition,
\renewcommand{\thefootnote}{}
\footnote{*Corresponding author: Hao Chen}
\end{abstract}    
\section{Introduction}
\label{sec:intro}

\begin{figure}[t]
  \centering
  \includegraphics[width=\linewidth]{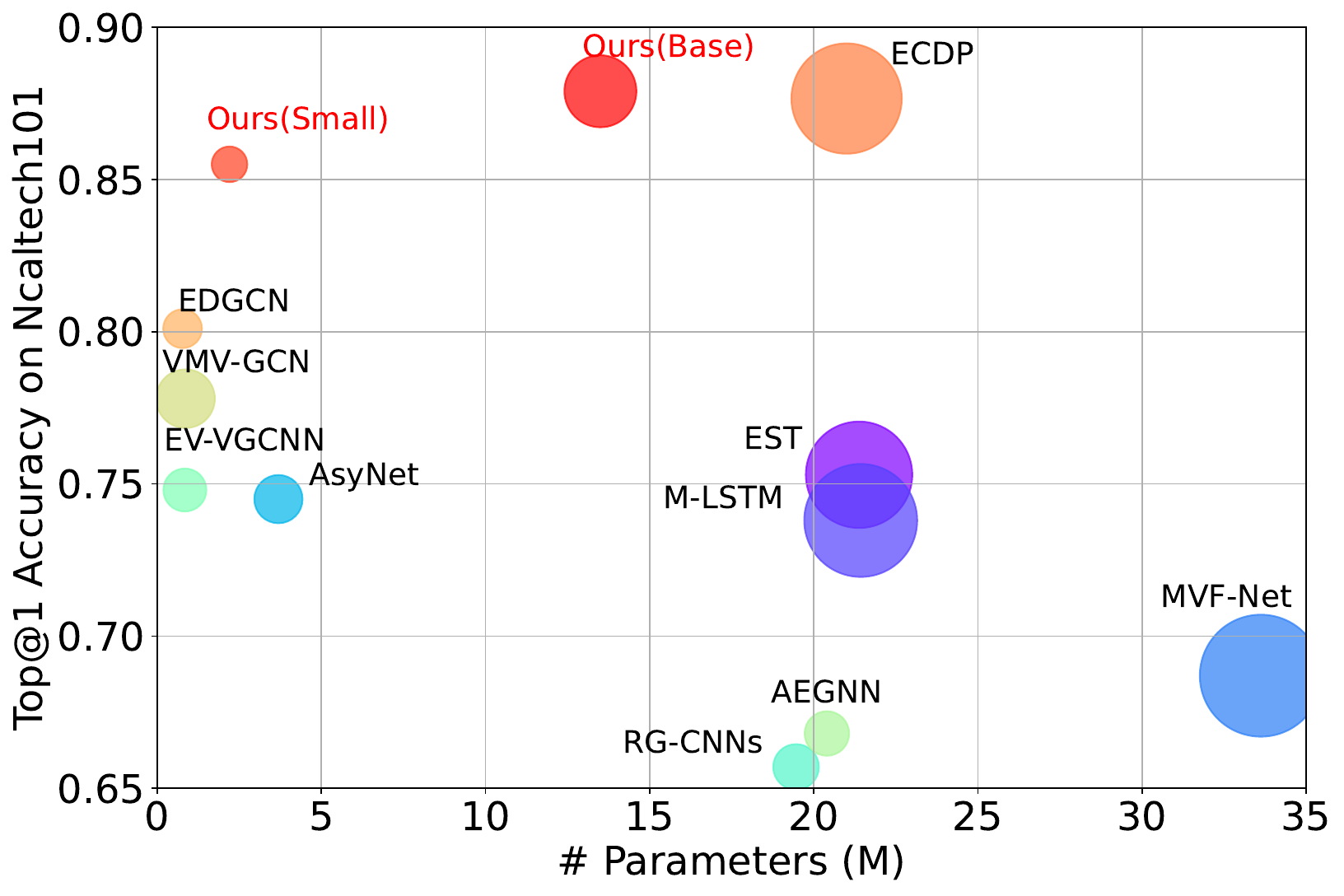}

   \caption{Comparison to state-of-the-art methods on N-caltech101 dataset in terms of accuracy and complexity. \textbf{FLOPS is proportional to the size of the circle associated with the model.}}
   \label{fig:complexity}
\end{figure}

The event camera \cite{brandli2014240, lichtsteinerposch}, a biomimetic sensor, operates by asynchronously reporting an event point when the brightness change of a single pixel surpasses a predetermined threshold. Its sparse and non-redundant stream output allows for recording sparse illumination changes with high temporal resolution and dynamic range. In comparison to traditional cameras, the event camera offers benefits including low latency, high dynamic range, and low power consumption. Therefore, the event camera is increasingly being applied in the field of computer vision, such as in denoising \cite{wang2020event}, semantic segmentation \cite{evsegnet, ess}, high frame rate video reconstruction \cite{rebecq2019high}, HDR \cite{han2020neuromorphic}, and so on.

 % It performs better in low dynamic range environments, such as at night or in mixed light scenes like tunnels, and avoids motion blur in high-speed moving scenes. 
% It excels in low dynamic range environments, such as at night or in mixed light scenes like tunnels, and mitigates motion blur in high-speed moving scenes.

As an emerging field, event cameras often suffer from a lack of sufficient labeled data, which becomes one of the main bottlenecks limiting the model capability.
In light of the notable achievements of self-supervised learning (SSL) in conventional modalities, \cite{mem,yang2023event} apply SSL methods from traditional image modalities to the event domain by converting the event sequence into 2D images. Nevertheless, this approach significantly undermines the core strengths of event cameras in practical applications, which lie in their capacity to capture high-speed temporal information and leverage the sparsity of data.

What we truly need is a self-supervised model specifically designed for event data, capable of preserving its valuable temporal information while possessing strong generalization capabilities and practical applicability (\eg, Figure \ref{fig:complexity}). 
To achieve this, we move away from the previous paradigm of transforming event sequences into 2D images for SSL. Instead, we aim to pre-train a voxel-based backbone. Compared to frame-based 2D representations, voxel-based representations have large advantages in retaining the temporal motion cues and data sparsity for lightweight implementation.
% motivation of designs in the encoder 

Therefore, in this paper, we propose to use the masked modeling idea to pre-train a voxel-based event model. Performing masked modeling on event voxels has two distinct challenges over doing on the traditional image domain: 1) deficiency of large-scale high-quality realistic datasets for pre-training. Current event datasets are generated through either rapidly shaking the camera in front of a screen with diverse trajectories and speeds, or using a mobile device equipped with an event camera operating in constrained scenarios. This results in substantial cross-dataset variance and a limited number of samples. 2) data sparsity and non-uniformity, as event information is only generated at locations where motion occurs. Hence, a data-efficient mask modeling method for event data is in demand.
% Existing event datasets are recorded by either quickly shaking the camera in front of the screen with significantly varying trajectories and speeds, or using a mobile device equipped with an event camera running in limited scenarios, thus holding large cross-dataset variance and limited samples.

\begin{figure}
  \centering
  \begin{subfigure}{0.325\linewidth}
    \includegraphics[width=\linewidth]{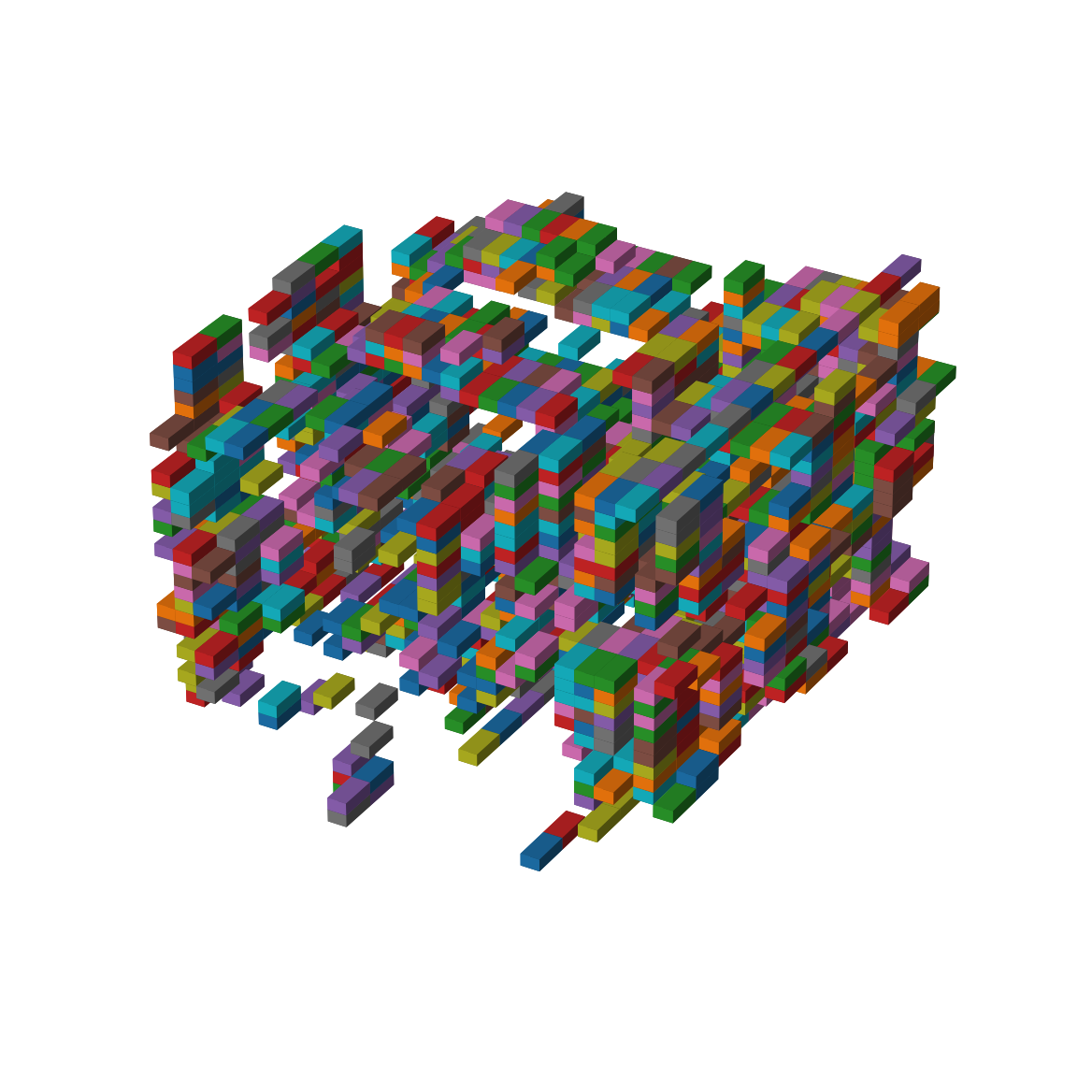}
    \vspace{-1cm}
    \caption{Origin.}
    \label{fig:short-a}
  \end{subfigure}
  \hfill
  \begin{subfigure}{0.325\linewidth}
    \includegraphics[width=\linewidth]{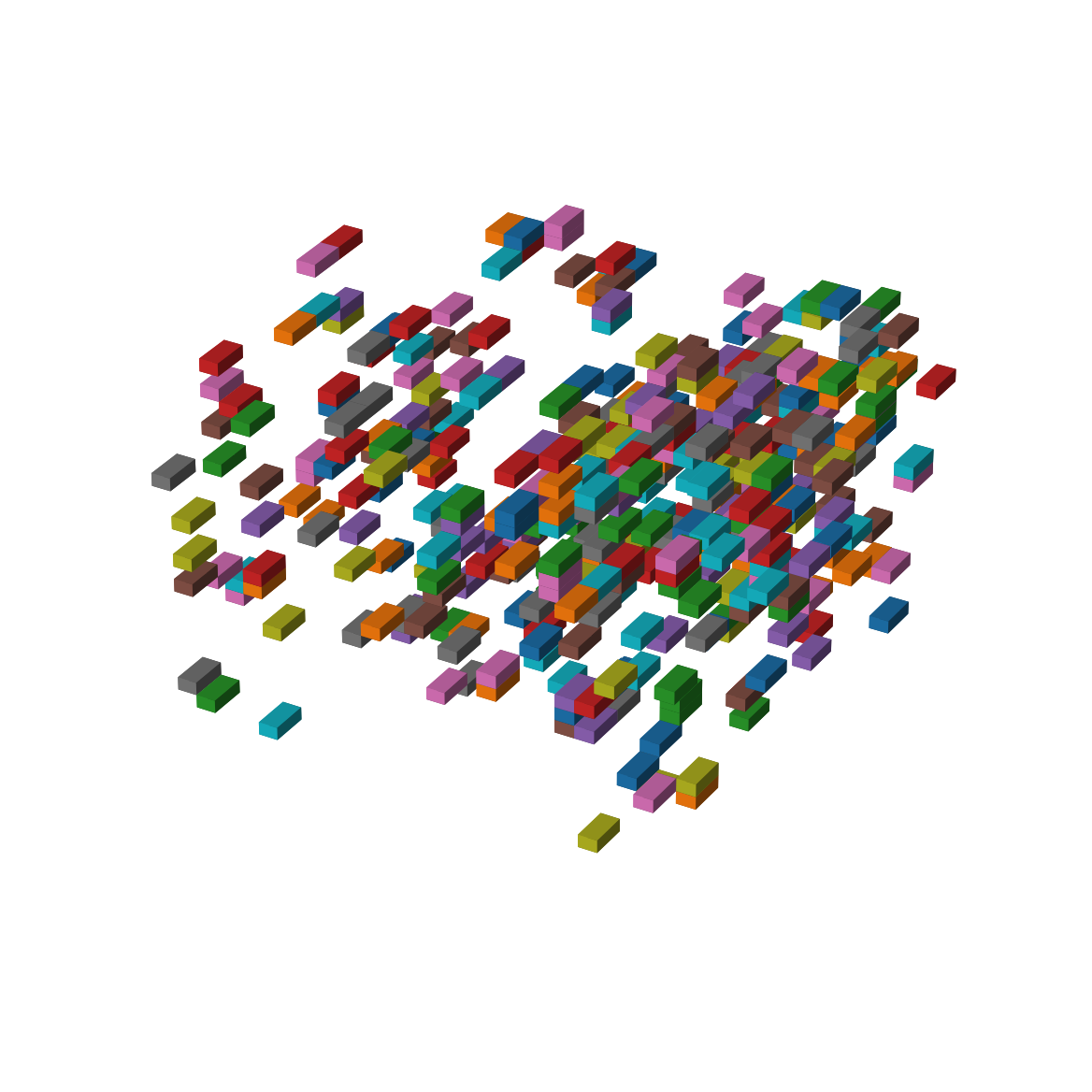}
    \vspace{-1cm}
    \caption{Random mask.}
    \label{fig:short-b}
  \end{subfigure}
  \hfill
  \begin{subfigure}{0.325\linewidth}
    \includegraphics[width=\linewidth]{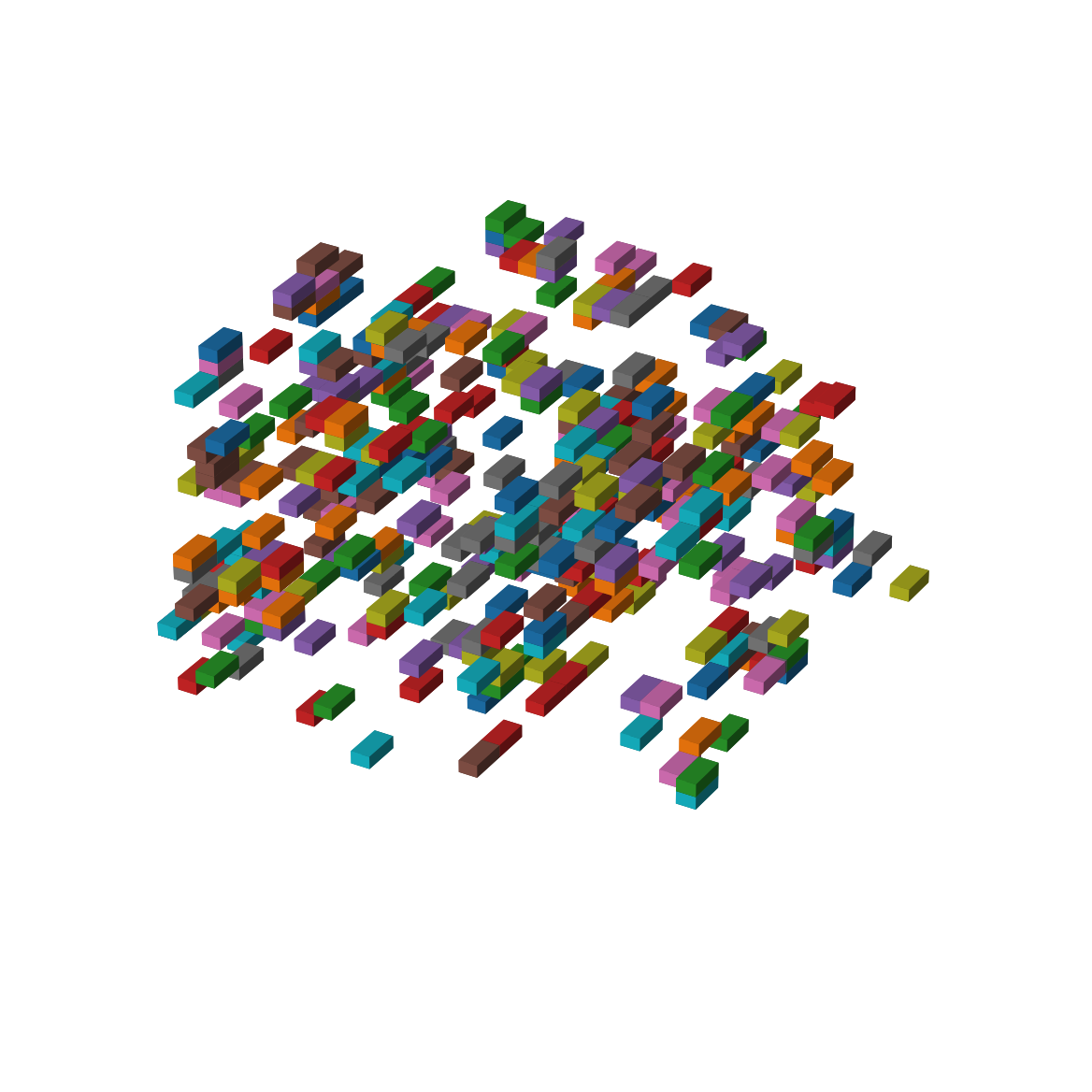}
    \vspace{-1cm}
    \caption{Uniform mask.}
    \label{fig:short-c}
  \end{subfigure}
  \vspace{-0.5cm}
  \caption{Visualization. Masked voxels are dropped. (a) shows the raw voxel input. (b) is the visible voxels after global random masking. Dense regions will be recovered more easily. (c) is the visible voxel after semantic-uniform masking. It balances the learning difficulty of each local semantic.}
  \label{fig:short}
\end{figure}

To achieve this goal, our first strategy is to tailor a masking method that specifically addresses the sparsity and non-uniformity of event data. Because the global random masking method \cite{he2022masked} does not consider such uneven distribution, it leads to varying difficulties in reconstruction for different regions (see Figure \ref{fig:short-b}). Specifically, denser areas with more visible tokens are easier to reconstruct while sparser areas are more challenging. To balance the difficulty of reconstruction and encourage the encoder to fully learn each semantic part, we propose a \textbf{Semantic-uniform Masking} approach that clusters the voxels into several parts with different local semantics and applies the same masking proportion for each cluster (see Figure \ref{fig:short-c}). This allows every region of each sample to contribute to the reconstruction, enabling our encoder to learn rich local semantics and comprehensive global understanding even with limited data.

% Due to the sparsity and non-uniformity of event data, the previous global random masking strategy \cite{he2022masked} is with high risk of masking or unmasking all voxels in a local part, leading to the failure of learning their representations. To address this issue, we propose a new semantic-uniform masking method. This approach involves clustering the events and performing masking within each cluster. By doing so, we ensure a more balanced and representative masking of the data, allowing for learning for each region and more completed contextual information.
% and 
Another method we proposed to achieve data-efficient pre-training is to focus on the reconstruction task. We explicitly disentangle the traditional hybrid global MAE idea \cite{he2022masked} into two parallel branches to reconstruct spatio-temporal local details and cross-region global semantics, respectively. In the local reconstruction branch, the masked voxels in each cluster are reconstructed by visible voxels in this cluster. This branch is designed to explore intra-cluster local correlations. A parallel global semantic reconstruction branch, which performs cluster-level masking and generates high-level representations of masked clusters by the high-level features from other visible clusters, is designed to capture global understanding and inter-cluster correlations. With two branches working complementarily, their shared encoder is endowed with capturing global semantics and low-level statistics. Compared to the hybrid MAE approach, we simplify the learning process by decomposing the reconstruction task, allowing for easier and faster learning of multi-scale representations with a small amount of pre-training data.

In summary, our contribution is as follows:

1.	In the event camera domain, this work is the first SSL method specifically designed for a spatio-temporal voxel-based backbone, and we do not rely on paired RGB data.

2.	We propose a disentangled masked modeling idea that can effectively reduce the learning difficulty under low data volume and improve SSL performance. 

3. We propose a semantic-uniform masking method to enable unbiased pre-training for each region and learning of completed global semantics.

4.	Our pre-trained model is lightweight and holds a strong generalization ability. It consistently outperforms state-of-the-art models by a significant margin across a wide range of tasks. 

%-------------------------------------------------------------------------

\section{Related work}
\label{sec:Related work}

%-------------------------------------------------------------------------
% \subsection{Self-supervised learning}
% Self-supervised learning is to explore the inborn supervision signals rather than relying on the human-annotated labels to train a deep model. Early SSL works focus on designing various pretext tasks such as image colorization, jagsaw puzzles and rotation prediction. However, the learned representations are typically biased to specific low-level properties. Nowadays, contrastive learning and masked modeling have been the two dominant lines in the SSL community for their great capability in learning high-level semantics.
% \subsubsection{Contrastive learning}
% Contrastive learning encourages the encoder to learn discriminative high-level semantic features by minimizing the distance between positive samples and maximizing the distance between negative samples in the high-dimensional feature space. Early contrastive learning methods \cite{} use fixed sets of negative samples, limiting the diversity of learned features. To address this issue, MoCo introduced a momentum update mechanism to enhance the diversity of negative samples. SwAV proposes a clustering-based approach and BYOL introduces an online prediction network for contrasting to overcome the reliance on generating negative samples. However, contrastive learning methods  highly rely on data augmentation and mainly focus on high-level semantic feature learning, neglecting lower-level features.

\subsection{Visual masked modeling}
Masked modeling is a simple pipeline that learns effective representations through mask reconstruction, which has been popularly used in image \cite{he2022masked, xie2022simmim}, video \cite{tong2022videomae,wang2023videomae,wei2022masked}, and point cloud \cite{min2022voxel, yang2023gd, pang2022masked, zhang2022point}. 
% Related research has made tremendous progress. The event stream shares certain similarities with video and point cloud in terms of data attributes. 
Conventional masked modeling methods applied to these domains are not directly applicable to event data due to their unique characteristics. Event data differs from images in that it includes temporal attributes, sets itself apart from videos with higher temporal resolution, asynchronicity, and sparsity, and distinguishes itself from point clouds through its three-dimensional spatiotemporal nature and feature diversity.
% However, conventional methods applied to these domains are not applicable to event data, as it differs from images by holding temporal attributes, distinguishes from videos by holding higher temporal resolution, asynchronicity, and sparsity, and differs from point clouds in its three-dimensional spatio-temporal nature and feature diversity.
 % Similarly, event data can be conceptualized as event point clouds in three-dimensional space. 
 
%MAE is to reconstruct to ..
%Simmim
%Following this idea, other modalities, such as video, 3D point cloud…  
%% for point cloud
%PointMAE
%PointM2AE
%GD-MAE
%voxel-MAE (difference to its masking strategy)
%% for video
%VideoMAE, VideoMAE++
%These methods are not applicable to event domain, as event data holds distinctive properties from other visual modalities, such as....   

%For event data, transfer event sequences to 2D frame, MEM, ICCV23 (MEM, ICCV 以及NIPS21等其他具体任务的自监督实现)
% task-specific SSL
% Self-Supervised Learning of Event-Based Optical Flow with Spiking Neural Networks
% introdce a contrast maximization loss to improve the optical flow estimation accuracy.

% Back to Event Basics: Self-Supervised Learning of Image Reconstruction for Event Cameras via Photometric Constancy

% EventPoint: Self-Supervised Interest Point Detection and Description for Event-based Camera

% Pre-training 
\subsection{Self-supervised learning for event data}
Most existing SSL methods for event data are designed for a determined task using task-specific constraints, such as in optical flow estimation \cite{hagenaars2021self}, image reconstruction \cite{paredes2021back}, interest point detection and description \cite{huang2023eventpoint}, and video deraining \cite{wang2023unsupervised}. For pre-training event models, MEM \cite{mem} explores the feasibility of transferring the BEiT \cite{beit}-style SSL framework to event data by converting events to 2D images. ECDP \cite{yang2023event} introduces paired RGB images to perform intra-modal and inter-modal contrastive learning. However, they mainly follow the SSL approach in the image domain to train a frame-based model, which leads to the loss of motion information and limited generalization ability.
% b. By matching paired event data and RGB data for training, it exhibits superior performance on downstream tasks. 
% ECDP \cite{yang2023event} proposes a self-supervised contrastive learning framework and a series of event data augmentation methods. By matching paired event data and RGB data for training, it exhibits superior performance on downstream tasks. 
% \cite{li2022neuromorphic} introduces the neuromorphic data augmentation technique and attempts to demonstrate its effectiveness within the SimCLR \cite{chen2020simple} framework.
%\cite{wang2023unsupervised} proposed a cross-modal contrastive learning method to achieve unsupervised video deraining using paired video frames and event data.
% However, contrastive learning usually relies on powerful data augmentation to improve model performance the inherent sparsity of event data and the temporal motion information have not been fully explored.
\begin{figure*}[t]
\centering
\includegraphics[height=0.5\textwidth, width=0.999\textwidth]{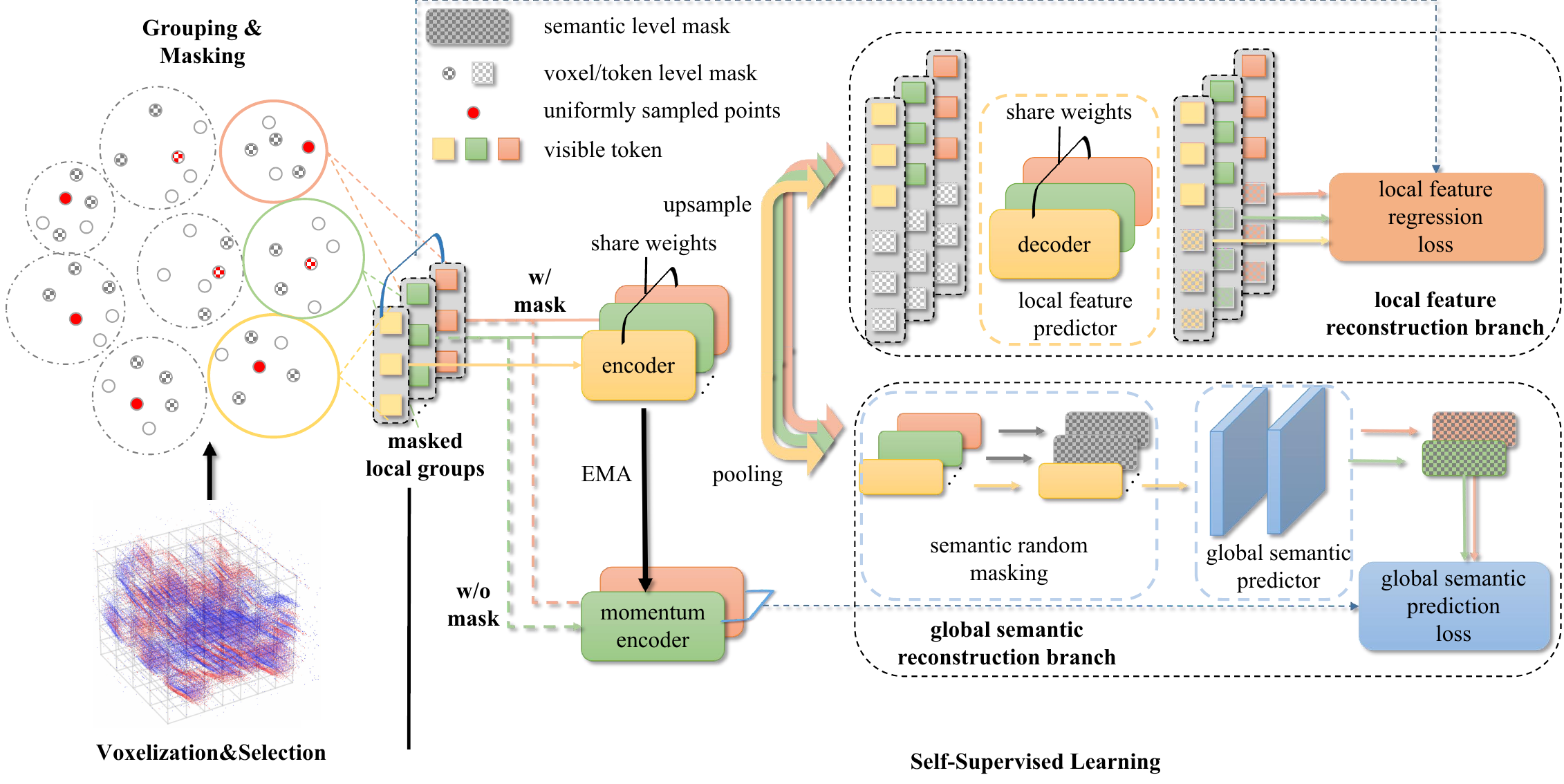}
\caption{Overview of our pre-training framework. 
% The left section depicts our data processing workflow. Initially, raw event data is voxelized and filtered based on the number of event points within each voxel. Subsequently, the most representative N voxel coordinates (\textcolor{red}{red points}) with
% their surrounding spatio-temporal structure are uniformly sampled from the voxels, and each local structure is randomly masked and then fed into the encoder respectively. Our self-supervised proxy tasks consist of two branches. (I) \textbf{Local Feature Regression Branch}: In the upper right section of the illustration, masked voxel feature reconstruction is performed within each local structure. (II) \textbf{Global Feature Prediction Branch}: In the lower right section of the illustration, each local structure generate summary tokens by encoder and a simple mean-pooling. Further masked semantic prediction is performed from a global view.
%Further random masking is applied at the summary token level, and another decoder is employed to complete the masked reconstruction of the summary tokens. 
%The reconstruction target is obtained using the momentum encoder, which is updated through Exponential Moving Average (EMA).
\textbf{Data processing workflow}: Left- Voxelizing and filtering raw event data. Then, Each uniformly sampled region is randomly masked and fed into the encoder separately. (I) \textbf{Local Feature Reconstruction Branch}: Upper right - Masked voxel feature reconstruction within each local structure. (II) \textbf{Global Semantic Reconstruction Branch}: Lower right - Summary tokens generated by encoder and mean-pooling for each region, followed by masked semantic prediction globally.
}
\label{learning architecture}
\vspace{-1.3em}
\end{figure*}

\noindent \textbf{Difference.} Our SSL holds several distinguished properties over previous event pre-training methods:
1. Voxel-based backbone rather than frame-based, thus retaining temporal motion cues and holding sparsity. 
2. Does not rely on paried RGB iamges.
3. Lightweight and fewer computational cost for better practical applicability.
4. Better performance and stronger generalization ability across a larger range of tasks.

\section{Method}

\subsection{Overview}
As illustrated in Figure \ref{learning architecture}, our pre-training framework mainly consists of four stages: 
1) Voxelization and selection. It aims to keep the sparsity and motion clues of event sequences efficiently. 
2) Grouping and Masking: group the voxels into clusters and perform uniform masking in each cluster to facilitate region-balanced pre-training.
3) A lightweight encoder, which is in charge of feature aggregation and capturing multi-scale spatio-temporal cues. 
4) Disentangled reconstruction, a local voxel-level reconstruction branch to reconstruct the masked voxels in each cluster given the visible voxels, and a global semantic reconstruction branch that predicts the high-level features of masked clusters by visible ones. 
% intra-cluster correlation and
% the inter-cluster correlation and
\subsection{Voxelization and selection}
Formally, event data $\{e_i\}_{i=1}^{n}$  with length of $n$ can be described as a sequence with four properties $ \{x_i, y_i, t_i, p_i\}_{i=1}^{n}$: the occurred location $(x_i, y_i)$, the triggered time stamp $t_i$ and  the polarity $p_i$, respectively.  

\noindent\textbf{Voxelization.} Voxelization \cite{deng2022voxel,deng2023dynamic} is an efficient way to reduce computational and storage costs while maintaining the sparsity, local relations, and temporal cues in event data. Given the voxel size $(v_w, v_h, v_t)$, the 3D space is evenly divided into voxels. For each voxel, the internal events are accumulated as voxel-level features: 
\begin{equation}
\mathcal{F}_i^{2 d}(x, y)=\sum_j^{N_v} p_j^{i n} \delta\left(x-x_j^{i n}, y-y_j^{i n}\right) t_j^{i n}
\label{Normalize}
\end{equation}
where $N_v$ is the number of internal points and $(x, y)$ is within the spatial range contained by the voxel.

\noindent\textbf{Selection.}
We adhere to the prior work \cite{deng2022voxel} to implement voxel selection. In detail, we select the top $N$ voxels $\mathcal{F}^{2 d} \in \mathbb{R}^{N \times v_w v_h}$ and corresponding coordinates with the highest count of internal events for input. It effectively preserves the global structure while reducing noise. 

\subsection{Semantic-uniform masking}
The conventional global random masking strategy \cite{he2022masked} is tailored for 2D dense image representations, and it applies masking indiscriminately to each region. However, event streams demonstrate sparse and non-uniform distribution in the $(x,y,t)$ 3D space. Consequently, the global random masking strategy can easily result in unbalanced reconstruction difficulties for different regions.
% the complete loss of certain sparse areas, leading to an incomplete overall semantic structure.
% without which may not adequately address the spatial variations in data distribution. 
To address this problem, we develop a straightforward yet effective local masking strategy to facilitate uniform learning of essential local semantics.
% within each segment and motion information across diverse segments of the three-dimensional sparse space.
As depicted in the top left of Figure \ref{learning architecture}, we partition the sample into several local spatio-temporal structures based on voxel coordinates and apply random masking within each local structure.
 % Therefore, we design a simple and effective local masking strategy to help the model evenly learn key semantics in each part and motion information across different parts of the three-dimensional sparse space. As illustrated in Figure \ref{learning architecture}, we partition the sample into several local spatio-temporal structures based on voxel coordinates and apply random masking within each local structure.
% \noindent\textbf{Masking}
Specifically, we gain the most representative $N$ voxel coordinates in the 3D space-time structure by uniformly sampling the input voxels. 
%(最远点采样？)
Each representative voxel searches for the $K$ nearest to its space-time coordinates. In doing so, the voxels are constructed into $N$ local parts $X=\left\{X_i\right\}_{i=1}^n, X_i \in \mathbb{R}^{K \times v_w v_h}$. 
With a mask ratio of $\rho_1$, each part is used as an independent input, and the internal voxel features $X_i$ are randomly masked respectively, and divided into visible part voxel features 
$X_i^v \in \mathbb{R}^{K_v \times v_w v_h}, K_v=\lfloor (1-\rho_{1})K \rfloor$
and 
masked part voxel features 
$X_i^m \in \mathbb{R}^{K_m \times v_w v_h}, K_m=K-K_v$.

% \begin{figure}
% \centering
% \includegraphics[height=0.28\textwidth, width=0.4\textwidth]{figures/model.png}
% % \includegraphics[width=0.9\linewidth]{Figs/Figure3.pdf}
% \caption{Our model architecture. }
% \label{model}
% \vspace{-1.3em}
% \end{figure}
% An efficient and versatile 
\subsection{Encoder}  
\label{encoder}
Events differ from traditional visual data (such as images and point clouds) by featuring mixed spatial and temporal coordinates, resulting in a distinct representation contrast.
Given this gap, we follow the previous work \cite{deng2022voxel} to perform local aggregation and insert it into the vanilla transformer architecture to search and aggregate neighbors according to each attribute to comprehensively consider the spatio-temporal relationships in the multi-head self-attention layer.

% implement small modifications  to the vanilla transformer architecture, specifically aggregating neighbor features based on spatio-temporal relationships in the multi-head self-attention layer. 
A down-sampling layer is introduced between each stage. From the multi-stage encoder, we obtain the $S$-stage representations $\{Y^v_{ij}\}_S^{j=1}$ for the visible voxels in region $X_i$.

\begin{table}
\centering
\begin{tabular}{ccccc}
\hline
Models     & Channels      & Layers & \#Params \\ \hline
Small       & {[}64,128,256,512{]} & {[}1,1,1,1{]} & 2.1M     \\
Base       & {[}96,192,384,768{]} & {[}2,2,2,2{]} & 13.4M    \\ \hline

\end{tabular}
\caption{The architecture hyper-parameters of our model variants}
\label{model variants}
\end{table}

\subsection{Local feature reconstruction branch}
% \noindent \textbf{Reconstruction} 
The voxel-level masked features $\Bar{X_i}^m$ are reconstructed by a vanilla transformer decoder $\text{Dec}_{L}$:
\begin{equation}
\Bar{X_i}^m = \text{Dec}_{L}(Y_i^v, T^{L}, R_i^v, R_i^m),
\end{equation}
where 
$R_i^v$
%\in \mathbb{R}^{(1-\rho_{1})K \times 3} $ 
and 
$R_i^m$
%\in \mathbb{R}^{\rho_{1}K \times 3} $ 
respectively represent the corresponding positional embeddings of visible and masked local voxels, $Y_i^v$ is upsampled from multi-stage representations 
$\{Y_{ij}^{v}\}_{j=1}^{S}$ by weighted interpolation referring to PointNet++ ~\cite{qi2017pointnet++}, and $T^{L}$ are learnable mask tokens representing masked voxels to be predicted. The objective function for the local feature reconstruction branch is:

\begin{equation}
\mathcal{L}_{local}=
\frac{1}{N\cdot K_m}
\sum_{i=1}^{N} 
%\sum_{j \in \mathcal{M}_1}
\left\|X_{i}^m-\Bar{X}_{i}^m\right\|_2^2 .
\end{equation}

\begin{table*}
  \centering
  \fontsize{9pt}{10pt} \selectfont
  \begin{tabular*}{\hsize}{@{}@{\extracolsep{\fill}}lcccccccc@{}}
\toprule        
Method   & Type$^{\ddagger}$ & \#Params &GFLOPs &Time  & Pre-Training Dataset & N-Cal          & N-C            & CIF10          \\ \midrule
\multicolumn{8}{c}{Supervised Pre-Training on ImageNet}                   \\ \midrule
EST \cite{gehrig2019end}               & F & 21.38 M  & 4.28   & 6.41 ms  & ImageNet &83.7 &92.5 &74.9 \\
M-LSTM \cite{Cannici_2020_ECCV}        & F & 21.43 M  & 4.82   & 10.89 ms & ImageNet &85.7 &95.7 &73.0 \\
MVF-Net \cite{mvfnet}                  & F & 33.62 M  & 5.62   & 10.09 ms & ImageNet &87.1 &96.8 &76.2 \\
ViT(ViT-S/16) \cite{yang2023event,vit} & F & 21.6 M     & 4.60   &-         & ImageNet &85.0 &96.8 &76.1 \\
ViT(ViT-B/16) \cite{yang2023event,vit} & F & 85.6 M     & 17.60  &-         & ImageNet &86.5 &\textbf{97.6} &77.5 \\

\midrule
\multicolumn{8}{c}{From Scratch}                                         \\ \midrule
%EST \cite{gehrig2019end}           & F & 21.38 M  & 4.28 & 6.41 ms  &  & 0.753          & 0.919          & 0.634          \\
%M-LSTM \cite{Cannici_2020_ECCV}        & F & 21.43 M  & 4.82 & 10.89 ms &  & 0.738          & 0.927          & 0.631          \\
%MVF-Net \cite{mvfnet}       & F & 33.62 M  & 5.62 & 10.09 ms &  & 0.687          & 0.927          & 0.599          \\
AsyNet \cite{messikommer2020event}      & F & 3.69 M   & 0.88 & -        &  & 74.5          & 94.4          & 66.3          \\
%EventNet \cite{Sekikawa_2019_CVPR}      & P & 2.81 M   & 0.91 & 3.35 ms  &  & 42.5          & 17.1          & 17.1          \\
RG-CNNs \cite{Graph-based}              & P & 19.46 M  & 0.79 & -        &  & 65.7          & 91.4          & 54.0          \\
EvS-S \cite{Li21iccv}                   & P & -  & - & -        &  & 76.1          & 93.1          & 68.0          \\
EV-VGCNN \cite{deng2022voxel}           & P & 0.84 M   & 0.70 & 7.12 ms  &  & 74.8          & 95.3          & 67.0          \\
AEGNN \cite{Schaefer22cvpr}             & P & 20.4 M   & 0.75 & -        &  & 66.8          & 94.5          &   .           \\
VMV-GCN \cite{Xie_2022_RAL}             & P & 0.86 M   & 1.30 & 6.27 ms  &  & 77.8          & 93.2          & 69.0          \\
%EV-Transformer \cite{li2022event} & P & 15.87 M  & 0.51 &-         &  & 0.789  .      & 0.954  .      & 0.709          \\
EDGCN \cite{deng2023dynamic}            & P & 0.77 M   & 0.57 & 3.84 ms  &  & 80.1          & 95.8          & 71.6          \\
GET \cite{peng2023get}                  & F & 4.5 M    & 3.10 & 17.55 ms &  & - & 96.7 & 78.1          \\%79.3
Ours (Small)     & P & 2.2 M    & 0.48 & 5.80 ms  &  & 81.4          & 96.4          & 71.0         \\
Ours (Base)     & P & 13.5 M   & 1.96 & 10.27 ms &  & 83.4          & 95.9          & 71.9          \\ 
\midrule
\multicolumn{8}{c}{Self-Supervised Pre-Training}                          \\ 
\midrule

%MAE (ViT-B/16) \cite{yang2023event, he2022masked}  & F   & 21 M     & 4.60 & -   & N-ImageNet & 0.677        & 0.953          & 0.687    \\
%MEM (ViT-B/16) \cite{klenk2022masked}        & F    & 86 M     & 17.60  & -   & Corresponding Dataset& 0.856          & \textbf{0.986} & -    \\
%ECDP(ResNet50) & F+I  & 23 M     & 4.10      & N-ImageNet & 0.871          &  0.980         & 0.748    \\
%ECDP(ViT-S)    & F+I  & 21 M     & 4.60      & N-ImageNet & 0.877          & 0.979          & 0.780    \\
ECDP \cite{yang2023event}  & F+I   & 21.6 M     & 4.60 & -   & N-Cal & 85.4        & 95.0   & 76.9    \\
%Ours(ViT-S) ${\dagger}$ \cite{yang2023event}  & P   & 21 M     & 83.20 & -   & N-Cal & 85.5        & 97.3   & 76.5    \\
%MAE(Base)      & P    & 13.5 M   & 1.96     & N-Cal (80\%) & 0.863              & 0.969          & 0.764          \\
Ours (Small)     & P    & 2.2 M    & 0.48 & 5.80 ms     & N-Cal & 86.0   & 97.1 & 75.9          \\
Ours (Base)      & P    & 13.5 M   & 1.96 & 10.27 ms    & N-Cal & \textbf{88.0}   & 97.1 & \textbf{78.6} \\

\bottomrule
  \end{tabular*}
  \caption{Comparison with state-of-the-art object recognition methods in terms of accuracy, model complexity (\#Params), and the number of FLOPs. 
  % ${\dagger}$: We reproduce self-supervised pre-training in ECDP style using N-Cal dataset and fine-tune it on three object recognition datasets. 
  $^{\ddagger}$:F: frame-based method; P: point-based method; F+I: frame-based method with extra paired RGB image data during self-supervised pre-training; '-' indicates that either the result is not reported or the source code is not publicly available. Best in bold. Following \cite{deng2023dynamic}, we calculate the complexity and FLOPs of the object classification model on the N-Cal dataset and measured the inference time on the N-C dataset using PyTorch on a Nvidia RTX 3090.}
  \label{tab:recognition}
\end{table*}

\subsection{Global semantic reconstruction branch}
This branch aims to enforce the encoder to efficiently construct the semantic correlations among all segments from a global view.
% Based on local reconstruction, global feature prediction branch aims to efficiently construct the connections from a global view. From the differences in the mask strategies employed for low-level and high-level features (i.e. random masking and block masking) as discussed in \cite{dong2022bootstrapped}, reconstruction on high-frequency details relies more on regions that are highly correlated with the mask voxel, i.e., near the voxel position. Cross-local connections are not obvious in high-frequency details, 
% , using high-level features as prediction targets to help guide the model to learn about high-level semantics
So we directly perform masking and reconstruction on the high-level semantic features. Concretely, we append a mean pooling layer to the output $\{Y_{iS}^{v}\}_{i=1}^{N}$ of the last stage of the encoder to obtain a collection of summary tokens $\{z_i\}_{i=1}^{N}$ that capture the essential information within their respective local contexts.
% \textbf{Masking}
%Consistent with the masking-predicting pattern of local branch, we also achieve this goal through a masking strategy. 
Then, we simply employ a random masking strategy with mask ratio $\rho_2$ to generate the index set $\mathcal{M}_G$ of masked summary tokens.
%The summary tokens is divided into two sets: visible tokens $Z^v=\{z_i^{v} | i \notin \mathcal{M}_2\}$ and masked tokens $Z^v=\{z_i^{v} | i \in \mathcal{M}_2\}$. 
%More specifically, the output pooling of each local voxel set, after being processed by the hierarchical encoder, is regarded as a high-dimensional vector $\{x_i\}{i=1}^n$ that retains the primary information within the local context. We randomly mask several local representations and partition them into $n_v$ visible local semantic representations $\{x_i^v\}{i=1}^{n_v}$ and $n_m$ masked portions, with the constraint that $n_v + n_m = n$. To facilitate this, we introduce additional learnable vectors as mask tokens, which serve to replace the semantic vector inputs of the masked portions within the decoder. This process yields corresponding outputs for the masked portions, denoted as $\{y_i^m\}_{i=1}^{n_m}$, allowing us to restore the original semantic features of the masked portions. The reconstruction loss function associated with the global representation masking can be formulated as:

% \textbf{Reconstruction} 
Given visible tokens $Z^v=\{z_i | i \notin \mathcal{M}_G\}$ as inputs, the masked summary tokens $\Bar{Z}^m =\{\Bar{z}_i | i \in \mathcal{M}_G\}$ predicted by $\text{Dec}_{G}$ can be represented as:
\begin{equation}
\Bar{Z}^m = \text{Dec}_{G}(Z^v, T^{G}, P^v, P^m),
\end{equation}
where $P^v$% \in \mathbb{R}^{(1-\rho_2)N\times 3} $ 
and 
$P^m$
%\in \mathbb{R}^{\rho_2N \times 3} $ 
respectively represent the corresponding positional embeddings of visible, masked local contexts, and $T^{G}$ are learnable mask tokens representing masked summary tokens to be predicted. 
We guide the decoder to complete the global feature prediction with the loss:
\begin{equation}
\mathcal{L}_{global}=\frac{1}{|\mathcal{M}_G|} \sum_{i\in \mathcal{M}_G}\left(1-\frac{\left\langle z_i, \Bar{z}_i\right\rangle}{\left\|z_i\right\|_2 \cdot \| \Bar{z}_i \|}_2\right),
\end{equation}
where the prediction targets $\{z_i | i \in \mathcal{M}_G\}$, represent the summary tokens obtained by the momentum encoder when fed with each complete local part 
$\{X_i | i \in \mathcal{M}_G\}$
as input.

% Further discussion will be provided in the experimental results section.

\subsection{Loss function}

%The loss function during the final model training process can be expressed as follows :
The final pre-training loss is 
\begin{equation}
\mathcal{L}_{total}=\mathcal{L}_{local} + \lambda \mathcal{L}_{global}, 
\label{total loss}
\end{equation}
where $\lambda$ is a hyperparameter to balance the contributions of the two reconstruction branches during pre-training and set to 1 without further tuning for all experiments.

\section{Experiment}

\begin{table*}[t!]
    \centering
    \small
    \begin{subtable}[t]{0.3\textwidth}
        \centering
          \begin{tabular}{@{}lcccc@{}}
\toprule        
Method              & N-Cal            &  \\ \midrule 
\multicolumn{2}{c}{From Scratch}             &  \\ \midrule 
YOLE  \cite{cannici2019asynchronous}      & 39.8                  &  \\
Asynet \cite{messikommer2020event}        & 64.3                  &  \\
NvS-S  \cite{Li21iccv}                    & 34.6                  &  \\
AEGNN  \cite{Schaefer22cvpr}              & 59.5                  &  \\
EDGCN   \cite{deng2023dynamic}            & 65.7                  &  \\
Ours(Small)          & 78.9                  &  \\
Ours(Base)          & 72.1                  &  \\ 
\midrule 
\multicolumn{2}{c}{Self-Supervised Pre-Training} &  \\ \midrule 
%MAE(base)          & 0.821         &  \\
Ours (Small)         & 81.1                  &  \\
Ours (Base)          & \textbf{83.3}                  &  \\
\bottomrule
  \end{tabular}
        \caption{Detection performance (mAP). ECDP does not provide results and source code for object detection.}
        \label{tab:object detection}
    \end{subtable}
    \hfill
    \begin{subtable}[t]{0.3\textwidth}
        \centering
  \begin{tabular}{@{}lcc@{}}
        \toprule
        Method & DDD17 & DSEC \\
        \midrule
        \multicolumn{3}{c}{Segmentation Methods} \\
        \midrule 
        EV-SegNet \cite{evsegnet}    & 54.81 & 51.76\\
        ESS\textcolor{blue}{${\dagger}$} \cite{ess}     & 61.37 & 53.29\\
        Ours (Small) & 58.70 & 56.74 \\
        Ours (Base) & 59.48 & 56.92\\
%        \midrule 
%        \multicolumn{3}{c}{CNN-Based Self-Supervised Pre-Training Methods} \\
        \midrule
        \multicolumn{3}{c}{Supervised Pre-Training on ImageNet} \\
        \midrule        
        ViT(ViT-S/16) \cite{vit}  &54.12 &42.92 \\
        ViT(ViT-B/16) \cite{vit}  &54.06 &45.55  \\
        %ResNet50 &59.25 &58.50 \\
        \midrule
        \multicolumn{3}{c}{Self-Supervised Pre-Training} \\
        \midrule
        %MoCo-v3 \cite{mocov3}   & 53.65 & 49.21\\
        %BeiT \cite{beit}  & 52.39 & 46.52\\
        %IBoT \cite{ibot}  & 49.94 & 42.53\\
        %MAE \cite{mae}  & 52.36 & 47.56\\
        ECDP\textcolor{blue}{${\dagger}$} \cite{yang2023event}  & 54.66 & 47.91\\
        Ours (Small) & 58.72 & 57.65\\
        Ours (Base) & \textbf{60.59} & \textbf{58.78}\\
        \bottomrule
  \end{tabular}
         \caption{Semantic segmentation (mIoU)
}
         %ESS utilizes additional image datasets and semantic labels during the training phase. 
        %ECDP requires paired RGB data during self-supervised pre-training.}
        \label{tab:semantic segmentation}
    \end{subtable}
    \hfill
    \begin{subtable}[t]{0.3\textwidth}
        \centering
        \begin{tabular}{@{}lc@{}}
\toprule        
Method  & DVS128 
\\ \midrule 
\multicolumn{2}{c}{From Scratch}
\\ \midrule
% EST \cite{gehrig2019end} & 0.941  & 4.28      & 21.38 M              \\ 
% MVF-Net  \cite{mvfnet}  &  0.950 & 5.62        & 33.62 M        \\

LIAF-Net \cite{LIAF-Net}                 & 97.6  \\
TA-SNN \cite{yao2021temporal}            & \textbf{98.6}  \\ 
% \multicolumn{3}{c}{Self-Supervised Pre-Training} \\
RG-CNN (Res.3D) \cite{Graph-based}       & 97.2  \\ 
EV-VGCNN \cite{deng2022voxel}            & 95.9  \\
VMV-GCN \cite{Xie_2022_RAL}              & 97.5  \\
EDGCN \cite{deng2023dynamic}             & 98.5         \\
GET \cite{peng2023get}                   & 97.9  \\
Ours(Small)             & 98.1  \\
Ours(Base)          & 98.1  \\
\midrule 
\multicolumn{2}{c}{Self-Supervised Pre-Training}
\\ \midrule
ECDP\textcolor{blue}{${\dagger}$}
\cite{yang2023event}             & 59.5  \\
%ECDP       & F\&I      & 0.591 & - & - \\
%MAE-Base       & P      & 0.985  \\
Ours(Small)               & \textbf{99.2}  \\
Ours(Base)                & \textbf{99.2}  \\
\bottomrule
  \end{tabular}
           \caption{Action recognition accuracy. %$^{\dagger}$: We use self-supervised pretrained ECDP model on the N-Cal dataset and fine-tune it on the DVS128 dataset.
        }
        \label{tab:action recognition}
    \end{subtable}

    \caption{Comparison to SOTA on object detection, semantic segmentation, and action recognition.          \textcolor{blue}{${\dagger}$} : model that requires paired RGB data during training.}
    \label{tab:downstream tasks}
\end{table*}

\subsection{Experimental setup}

\textbf{Choices of pre-training dataset.}
Due to its extremely short recording time (i.e. $50~\mu s$) per sample, N-ImageNet \cite{nimagnet} lacks sufficient motion cues for downstream tasks, particularly in action recognition that requires longer durations (e.g., $6000~ms$ in DVS128 Gesture Dataset \cite{dvs128}). Consequently, previous self-supervised methods \cite{yang2023event} primarily focus on spatial cues when using N-ImageNet as the pre-training dataset.
To overcome this limitation, we propose utilizing N-Caltech101 \cite{ncaltech} as our pre-training dataset instead. N-Caltech101 is recorded using an event camera with the RGB version of the Caltech101 image dataset. It comprises 8246 samples, each lasting $300~ms$, and covers 101 object categories. Although N-Caltech101 has simpler categories and fewer samples compared to N-ImageNet, it offers a more robust representation of motion cues and will benefit a larger range of downstream tasks.

\noindent \textbf{Implementation.}
For a fair comparison, We use the same train-test splits adopted in ECDP \cite{yang2023event} for N-Caltech101 (N-Cal) and CIFAR-10-DVS \cite{CIFAR10DVS} (CIF10) during pre-training and fine-tuning. In the self-supervised pre-training phase, the model is trained for 700 training epochs with a batch size set to 64. Voxel size $(v_w, v_h, v_t)$ is set to (5, 5, 25 ms) following \cite{deng2023dynamic}. 
We use AdamW and a cosine schedule \cite{Loshchilov_Hutter_2016} with a single cycle where we warm up the learning rate for 40 epochs to 3e-4. 
A complete input of 2048 voxels is divided into 16 parts by farthest point sampling, and the mask ratio is set to 80\%. 
%In the fine-tuning phase, the voxel size and batch size for the N-Cal sample remained unchanged, with a learning rate of 4e-3 and 200 epochs.%For semantic segmentation task, the fine-tune epochs are 100. 
 The detailed training settings are in the supplementary material.
Since the models to be compared are mainly divided into two categories, one pursues lightweight and efficient methods, such as EV-VGCNN \cite{deng2022voxel}, EDGCN \cite{deng2023dynamic}, etc. and the others use relatively heavy backbones from ViT~\cite{vit} families (\eg, ECDP \cite{yang2023event}),
%, MEM \cite{klenk2022masked}
%, etc. 
we build two variants to compare separately. The specific structure is shown in the Table~\ref{model variants}.

\noindent \textbf{Baselines.} 
\textit{1) Supervised learning methods.} The supervised baseline model primarily involves transferring the pre-trained model from the RGB domain on ImageNet to the event data domain.
\textit{2) Self-supervised learning methods.} For fair comparison, we perform the SOTA SSL method ECDP \cite{yang2023event} using the ViT-S backbone on the N-Cal dataset with the same pre-training epoch number as our method.  
Other settings during pre-training and fine-tuning of ECDP are consistent with the original paper.

\subsection{Performance on downstream tasks}
In this section, we evaluate our pre-trained model performance on various downstream tasks. 
%The number of voxels in each stage is set to $\{2048, 512, 128, 32\}$ for N-Cal and CIF10, $\{256, 128, 64, 32\}$ for N-Car (N-C), respectively.
\subsubsection{Object recognition}
The fine-tuning results are shown in Table \ref{tab:recognition}. Among the group ``training from scratch", our model achieves a very good balance in terms of parameter numbers, computational complexity, and accuracy, all of which are among the best in its class, demonstrating the efficacy and efficiency of our backbone. With our SSL pre-training, we obtain a large improvement by 4.6 points over training from scratch (88\% vs. 83.4\%), showing the effectiveness of our SSL scheme. Compared to the SSL method ECDP, which relies on paired RGB images to supply additional supervision signals for the event encoder, our self-supervised training method, without using paired RGB images, only holding its half parameters and GFLOPs, while outperforms ECDP by 2.6 points (88\% vs. 85.4\%). Compared to methods that have more parameters and are supervised pre-trained on ImageNet, our SSL model, with a lightweight backbone, outperforms them by a large margin (88\% vs. 86.5\%).

\subsubsection{Object detection}

% We utilize the N-Cal dataset to assess the model's performance in the object detection task, which demands both semantic understanding and simultaneous motion information capture capabilities. 

%Following the configuration in AEGNN \cite{Schaefer22cvpr}, we employ a YOLO-based object detection head and a weighted sum of class, bounding box offset, shape, and prediction confidence losses as the task loss. 
We implement YOLO for object detection, utilizing a task loss that combines class, bounding box, shape, and confidence losses, as specified in AEGNN \cite{Schaefer22cvpr}. 
%we finetune our detection model on N-Cal and the number of voxels in each stage is set to $\{2048, 512, 128, 32\}$. 
Table \ref{tab:object detection} shows that our pre-training scheme brings 11.2 mAP improvement over training from scratch (83.3\% vs. 72.1\%). It outperforms all supervised learning methods by a large margin.
% Our detection performance is evaluated using the eleven-point mean average precision (mAP) as the metric. 

\subsubsection{Semantic segmentation}
% Furthermore, we showcase the performance of our model in semantic segmentation on two datasets: 
We finetune our segmentation model on DSEC \cite{dsec, ess} and DDD17 \cite{ddd17, evsegnet}, respectively as done in other methods. 
%The number of voxels in each stage is set to $\{4096, 1024, 256, 64\}$. 
% We employ the mean Intersection over Union (mIoU) metric to evaluate the methods.

Semantic segmentation requires assigning labels at the pixel level, however, the sparse framework faces challenges in handling pixel-level tasks because it exclusively encodes the features of non-empty voxels. 
%Therefore, we simply build a lightweight decoder with 2-layers convolution blocks to generate dense queries from voxel grid representation and cross attention to aggregate voxel-level sparse features into patch-level dense features. 
Therefore, we simply implement a lightweight method to convert the sparse 3D representation into a dense 2D representation. After voxelization, non-empty voxels are retained, and the complete 3D space is treated as a multi-channel image to obtain $(v_w, v_h)$ sized patch embeddings as queries. The sparse features extracted from our backbone in 3D space are used as keys and values and are fed into the cross-attention layer to complete the conversion.
More details about the segmentation decoder architecture can be found in the supplementary material. The comparison results are reported in Table \ref{tab:semantic segmentation}. We observe a substantial improvement over ECDP (58.78\% vs. 47.91\%). Our method outperforms most other approaches by a significant margin, except ESS \cite{ess} on DDD17. It is important to note that ESS benefits from the use of additional paired RGB images and corresponding labels, which gives it an advantage.

\subsubsection{Action recognition}

% Action recognition tasks usually involve much longer time sequences, which puts forward higher requirements for long-term dependence on motion capture capabilities.
% When compared to self-supervised models that rely on two-dimensional input, the most significant difference is our superior performance in long-term sequence action recognition tasks.

% For a comprehensive comparison with previous self-supervised models, each action recognition sample is compressed into a two-dimensional image and then fed into the self-supervised ECDP model, resulting in underwhelming accuracy 59.5\%, which is not surprising, as the action recognition task is closely tied to the capacity to capture temporal actions, which is a capability previous single frame based self-supervised models lack.
% Based on the self-supervised learning framework we designed, our model's ability to capture temporal motion has been further improved and achieved the best results (98.6\% vs. 99.2\%). The strong transfer ability of the self-supervised model in long-term tasks is confirmed with a performance improvement from 98.1\% to 99.2\%.

We evaluate our self-supervised model on DVS128 \cite{dvs128}. 
%The number of voxels in each stage is set to $\{1024, 256, 64, 16\}$. 
From Table \ref{tab:action recognition}, we notice that the SOTA event SSL method ECDP, which transforms event sequences into 2D images for SSL, yields poor results (59.5\%) on action recognition. This indicates that the pre-trained model from ECDP has limited generalization ability. This is because the transformation of event sequences into 2D images results in a significant loss of temporal information, which happens to be crucial for action recognition. In contrast, our method can effectively capture both spatial and temporal information from the event inputs, leading to superior performance in action recognition. The transfer of our self-supervised model to action recognition also results in a noticeable improvement (98.1\% to 99.2\%).

\begin{table*}[t!]
    \centering
    \hfill
    \small
    \begin{subtable}[t]{0.31\textwidth}
        \centering
        \setlength{\tabcolsep}{2pt}
        \begin{tabular*}{0.9\textwidth}{@{\extracolsep{0.02pt}}ccc|c}
        Task        & N-Cal  & CIF10 & DSEC \\\hline
        MAE-Voxel      &86.3  &76.4 &58.4 \\
        Local Branch   &86.9  &78.0 &58.6 \\
        Dual Branches  &\textbf{88.0}  &\textbf{78.6} &\textbf{58.8}  \\
        \end{tabular*}
        \caption{
       \textbf{Different reconstruction objectives.}
       % We select MAE-style pre-training target as our baseline and progressively validate the significance of each branch.
        }
        \label{tab:proxy task}
    \end{subtable}
    \hfill
    \small
    \begin{subtable}[t]{0.38\textwidth}
        \centering
        \setlength{\tabcolsep}{3pt}
        \begin{tabular*}{0.9\textwidth}{@{\extracolsep{\fill}}ccccc|c}
        Strategy &Local &Global & N-Cal & CIF10 &DSEC \\\hline
        Random  & \checkmark &            & 86.9 & 77.6 & 58.0 \\
        Uniform & \checkmark &            & 86.9 & 77.5 & 58.6 \\
        Random  & \checkmark & \checkmark & 87.5 & 78.0 & 57.7 \\
        Uniform & \checkmark & \checkmark & \textbf{88.0} &\textbf{78.6} &\textbf{58.8}\\
        \end{tabular*}
        \caption{
        \textbf{Grouping strategy.} 
        Compare our uniform sampling to random sampling  for choosing $N$ most representative points, in our pre-training. `Local' and `Global' indicate whether to use the respective reconstruction branch.
        }
        \label{tab:grouping strategy}
    \end{subtable}
    \hfill
    
    \caption{
    Ablation study on classification (Accuracy on N-Cal and CIF10) and semantic segmentation (mIOU on DSEC).}
    \label{tab:ablation}
\end{table*}

% Ablation study on N-Cal, CIF10, and DSEC datasets. Accuracy and mIOU are employed to evaluate classification and semantic segmentation performance respectively.

% \subsection{Model Complexity}

\noindent \textbf{Result summary.} In the experiments for the aforementioned tasks, our self-supervised model consistently achieves the best performance by a significant margin, demonstrating the powerful generalization ability of our pre-trained model and the effectiveness of our pre-training method. Additionally, our method does not need paired RGB images, and has very few parameters and computational requirements, making it highly practical for real-world applications.

\subsection{Ablation study}
% In this subsection, we perform in-depth ablation studies on our core designs. 
We use our base backbone in the ablation experiments, and all settings are consistent with the default. We report the results on classification (N-Cal, CIF10) and segmentation (DSEC) to comprehensively investigate our designs.  
\subsubsection{Effectiveness of each reconstruction branch} 
Because there is currently no SSL method specifically designed for voxel-based event backbones, we craft a baseline by replacing the disentangled reconstruction method proposed in this paper with the hybrid global reconstruction method used in MAE \cite{he2022masked}, that is directly reconstructing all missing voxels using visible voxels. The voxelization and selection processes, as well as the encoder structure, remain unchanged. We refer to this strong baseline as \textbf{MAE-Voxel}.
% We explored two approaches to demonstrate the effectiveness of our solution:
% (1) Use uniform masks at the global level and pre-training in MAE style. 

\noindent \textbf{Local feature reconstruction.}
% The local uniform reconstruction strategy is one of our core strategies. 
Compared to MAE-Voxel in Table \ref{tab:proxy task}, our local reconstruction branch achieves noticeable improvement. 
We attribute it to that the locally uniform reconstruction strategy balances the reconstruction difficulty between sparse and dense parts, avoiding the encoder bias on dense regions. 
As a result, the local reconstruction branch enables the encoder to learn more completed local features. 
% This results in a more efficient encoding of the spatio-temporal features of each component.

\noindent  \textbf{Global semantic reconstruction.}
% Global semantic reconstruction is a fundamental design aimed at integrating local modeling into the global context. This approach involves adding a complete, natural global constraint for each independent local representation at high-dimensional semantic level, which enables more effective information interaction 
As shown in Table \ref{tab:proxy task}, additionally combining the global semantic reconstruction with the local semantic reconstruction branch (denoted by ``Dual Branches”) results in further improvement, demonstrating the effectiveness of the global semantic reconstruction. We think the underlying contribution of this branch is that it enforces the encoder to build correlations among local regions and learn global semantics. 
\subsubsection{Efficacy of disentangled masked modeling}
% \noindent \textbf{Efficient Event Data Learner}
Table \ref{tab:proxy task} has already demonstrated the effectiveness of each reconstruction branch designed by us, as well as the significant improvement brought by the combination of them. This validates the effectiveness of our disentangled reconstruction strategy. It decomposes the original hybrid reconstruction task into two parts: local details and global semantics, making it easier to comprehensively learn multi-scale representations. 
Next, we will verify the advantages of disentangled reconstruction in terms of training efficiency and data efficiency.

\begin{figure}
    \centering
    \includegraphics[width=\linewidth]{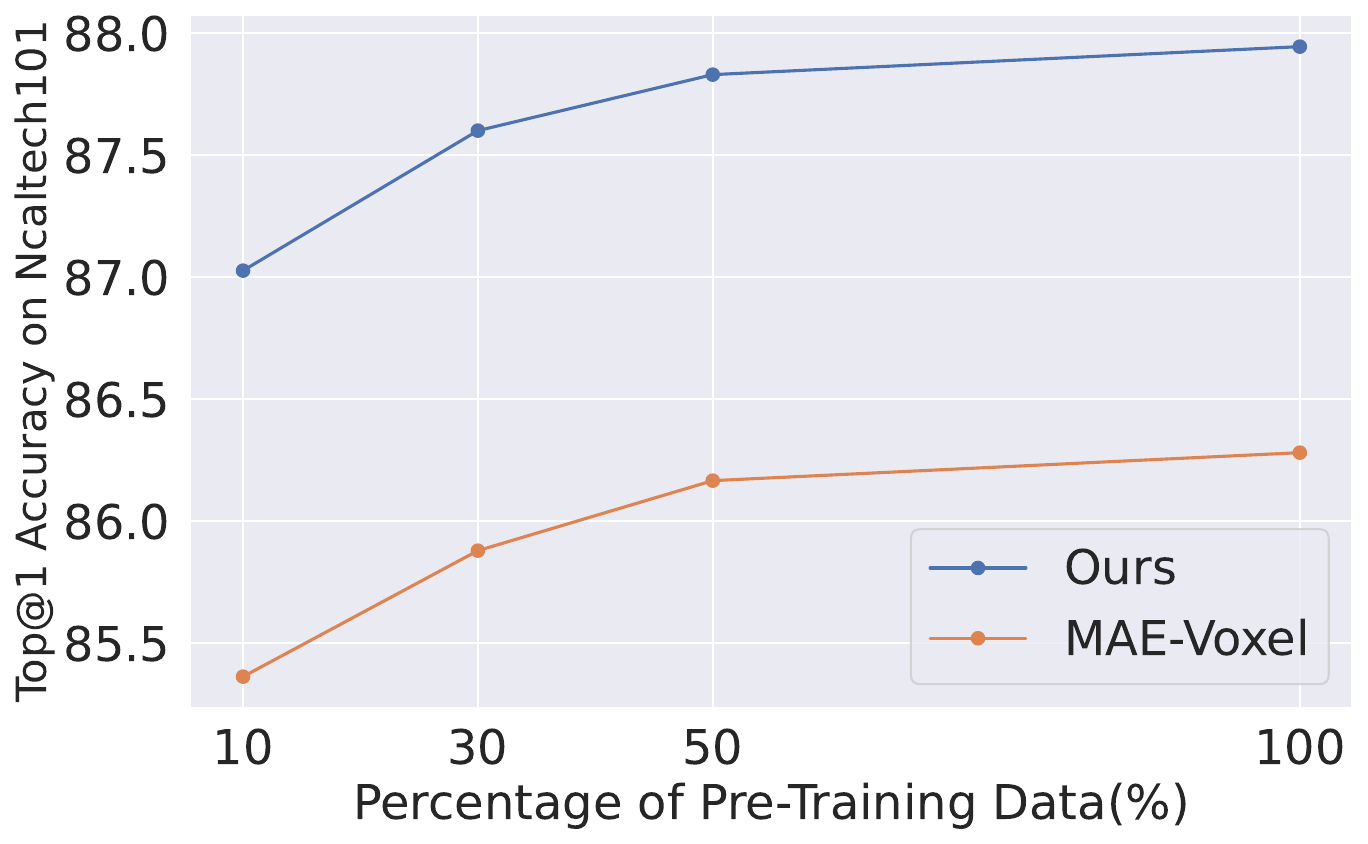}
    \caption{Different percentages of pre-training data.    }
    \label{fig:less_data}
\end{figure}

\noindent \textbf{Need less pre-training data.}
Our pre-training is data-efficient. To verify this, we choose subsets comprising 10\%, 30\%, and 50\% of the pre-training dataset, with each smaller subset being part of the larger subset. The number of training iterations for each subset is kept consistent with the original configuration. As shown in Figure \ref{fig:less_data}, even with reduced size of the pre-training dataset, our approach consistently maintains a significant advantage over the original hybrid MAE (\ie, MAE-Voxel), demonstrating its ability to learn representations more efficiently from limited pre-training data. With only 10\% pre-training data, our SSL method brings 3.6 points improvement over training from scratch (from 83.4\% to 87\%).

\begin{figure}
    \centering
    \includegraphics[width=\linewidth]{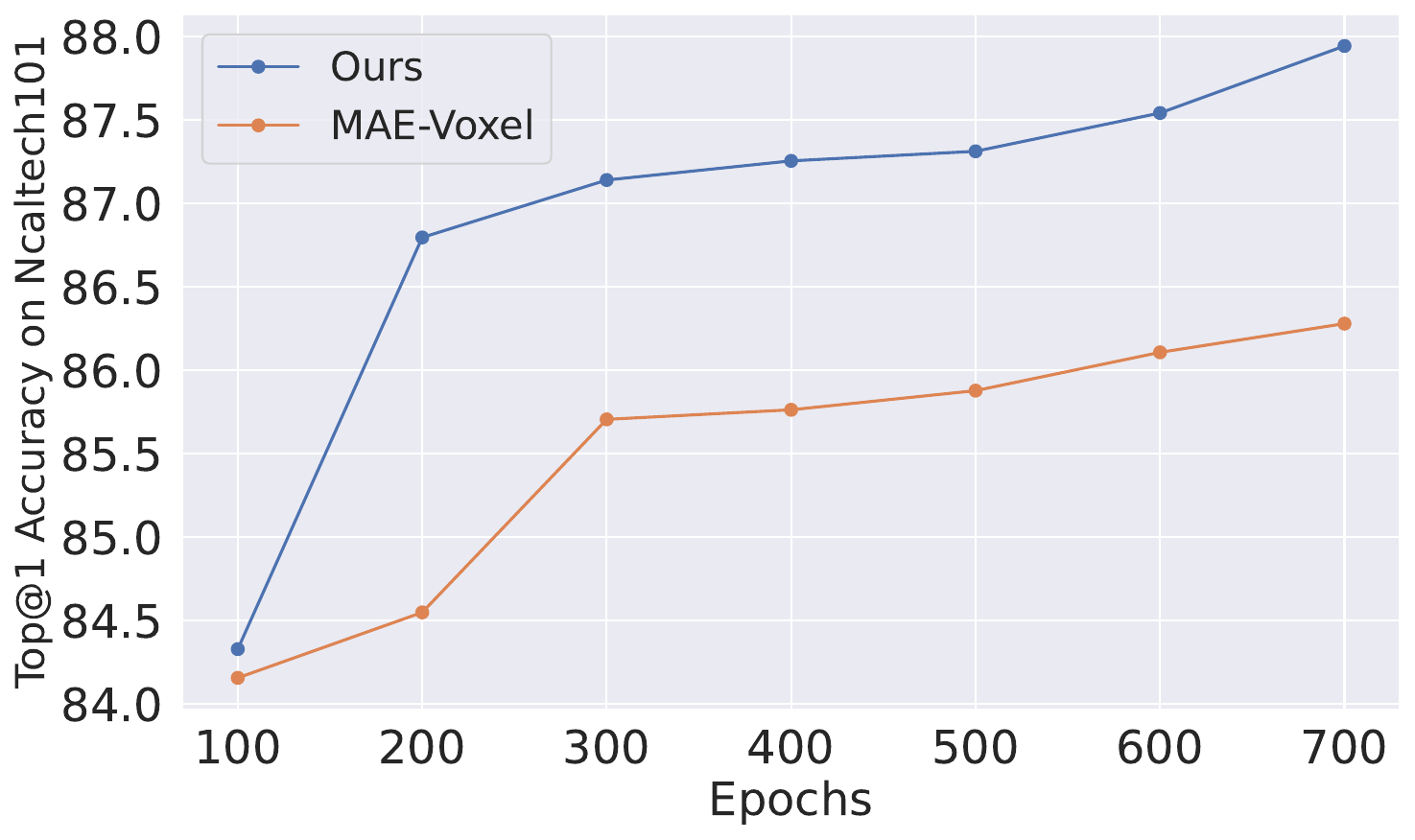}
    \caption{ Different pre-training epochs.     }
    \label{fig:lest_epoch}
\end{figure}

% \begin{figure}[thbp!]
%     \centering
%     \begin{minipage}[t]{0.49\linewidth}
%         \centering
%         \includegraphics[width=\linewidth]{figures/need_less_data.pdf}
%          \caption{
%          %Fine-tuning performance with varying pre-training data percentage. 
%          Different percentages of pre-training data.
%          }
%         \label{fig:less_data}
%     \end{minipage}
%     \begin{minipage}[t]{0.49\linewidth}
%     \centering
%     \includegraphics[width=\linewidth]{figures/need_less_epoch.pdf}
%     \caption{
%     %Fine-tuning performance with varying numbers of pre-training epochs
%     Different pre-training epochs.
%     }
%     \label{fig:lest_epoch}
%     \end{minipage}
%  \end{figure}

\noindent \textbf{Need less pre-training epoch.} 
We vary the number of pre-training epochs using our SSL method and MAE-Voxel, then perform transfer learning to object recognition. The comparison of fine-tuned results is shown in Figure \ref{fig:lest_epoch}. We consistently observe that our method outperforms MAE-Voxel across all pre-training epochs. Additionally, our pre-training method exhibits faster convergence speed, achieving high performance in just 200 epochs. This suggests the efficiency of our pre-training method and the success of our disentangled masked modeling approach in facilitating pre-training and enriching the pre-trained representations.
% can exceed the best results of the baseline method with only 200 epochs.

\subsubsection{Semantic-uniform masking} 
\begin{figure}
    \centering
    \includegraphics[width=\linewidth]{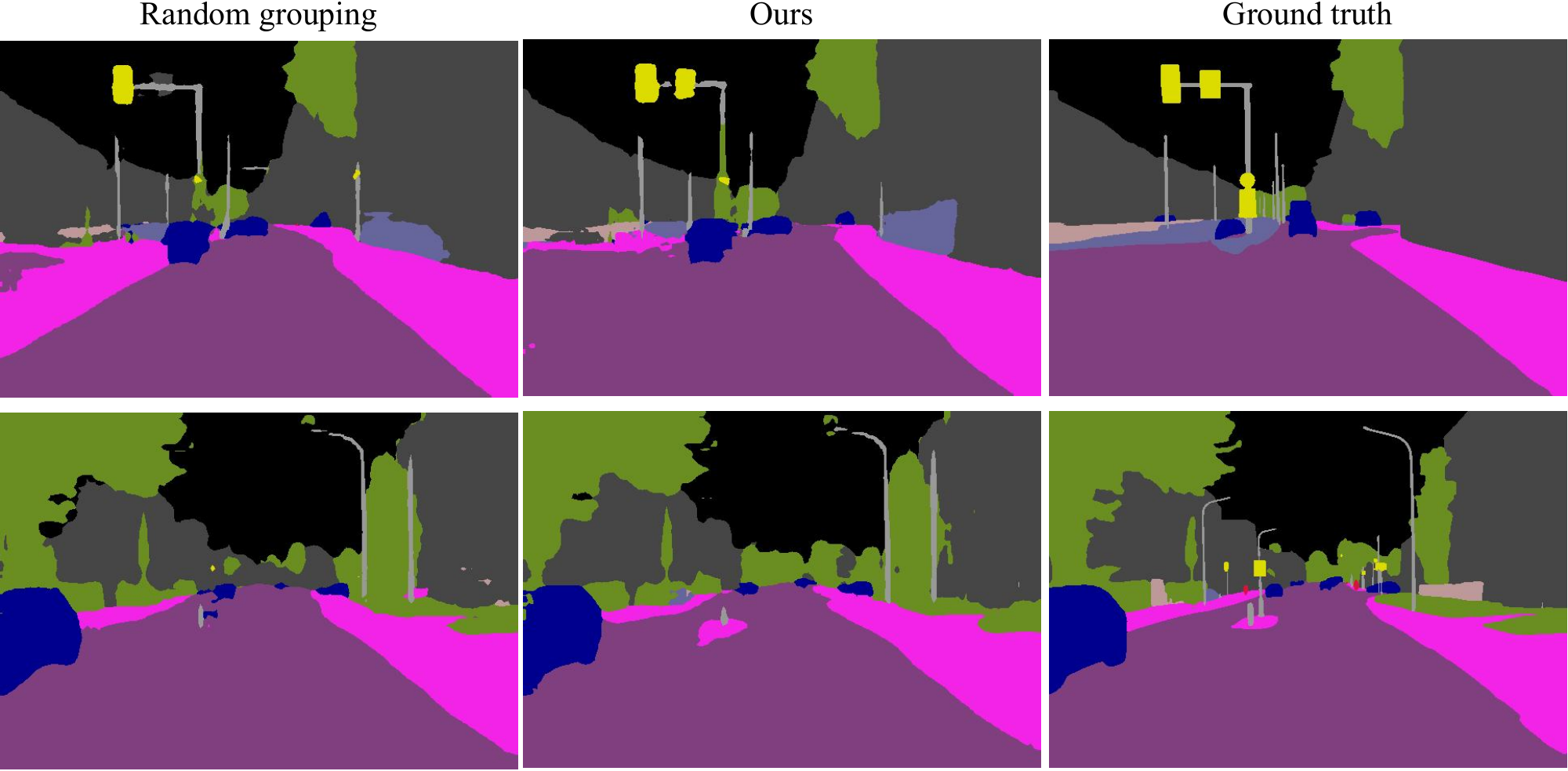}
    \caption{Qualitative comparison on the DSEC dataset. 
    % \textbf{In the first column,} the grouping strategy in our method is replaced with random sampling. \textbf{The second column} shows the predicted results of our model. \textbf{The last column} is the ground-truth segmentation image. The color of the pixel represents the segmentation category.
    }
    \label{fig:visualization}
\end{figure}

% Select local parts randomly rather than through uniform sampling.
% Both approaches result in performance degradation. It can be explained as (1) making the proxy task of establishing global connections change from reconstructing summary tokens to cross-local voxel-level features, and (2) preventing the proxy task of establishing connections from a complete global perspective, which confirms our effectiveness of our idea.
From Table \ref{tab:grouping strategy}, it can be observed that compared to random masking used in MAE \cite{he2022masked}, our uniform masking method shows improvements for both the local and global reconstruction branches. 
%The two decoupled branches are better able to focus on rich local voxel-level features and high-level global semantics.
This indicates that it allows the encoder to establish a more complete global structure, enabling the encoder to explore richer voxel features under different spatial and temporal distributions with more comprehensive global semantics. 
%avoiding the loss of some sparse yet key parts and 
%The two decoupled branches focus specifically on rich local voxel-level features and high-level global semantics.
It is worth noting that the uniform masking does not provide improvement for the local branch in the classification task but exhibits noticeable enhancement in the segmentation task. We think that this is because the loss of a small amount of local information has a minor impact on classification decisions, while the segmentation task requires more local information. Figure \ref{fig:visualization} shows that using our uniform masking method achieves more complete and accurate segmentation, especially for small regions.  

\section{Conclusion}
In this work, we propose a data-efficient voxel-based pre-training method for event data. We introduce the disentangled masked modeling method, which decomposes the original MAE into local feature reconstruction and global semantic reconstruction, thereby simplifying the reconstruction task and improving the sufficiency of pre-training. This approach accelerates the pre-training process and reduces the amount of required pre-training data. Additionally, we propose a semantic-uniform masking strategy to facilitate unbiased learning for each part. Our method outperforms existing state-of-the-art models across various tasks with large margins, which with very few parameters and computational complexity.
% Moreover, we present a more adaptive neighbor search selection method that considers the spatial-temporal information variations in different datasets. 
%{
%    \small
%    \bibliographystyle{ieeenat_fullname}
%    \bibliography{main}
%}
\appendix
\small
\bibliographystyle{ieeenat_fullname}
\bibliography{main}

\begin{thebibliography}{46}
\providecommand{\natexlab}[1]{#1}
\providecommand{\url}[1]{\texttt{#1}}
\expandafter\ifx\csname urlstyle\endcsname\relax
  \providecommand{\doi}[1]{doi: #1}\else
  \providecommand{\doi}{doi: \begingroup \urlstyle{rm}\Url}\fi

\bibitem[Alonso and Murillo(2019)]{evsegnet}
I{\~{n}}igo Alonso and Ana~C. Murillo.
\newblock Ev-segnet: Semantic segmentation for event-based cameras.
\newblock In \emph{CVPRW}, pages 1624--1633, 2019.

\bibitem[Amir et~al.(2017)Amir, Taba, Berg, Melano, McKinstry, Di~Nolfo, Nayak, Andreopoulos, Garreau, Mendoza, et~al.]{dvs128}
Arnon Amir, Brian Taba, David Berg, Timothy Melano, Jeffrey McKinstry, Carmelo Di~Nolfo, Tapan Nayak, Alexander Andreopoulos, Guillaume Garreau, Marcela Mendoza, et~al.
\newblock A low power, fully event-based gesture recognition system.
\newblock In \emph{CVPR}, pages 7243--7252, 2017.

\bibitem[Bao et~al.(2022)Bao, Dong, Piao, and Wei]{beit}
Hangbo Bao, Li Dong, Songhao Piao, and Furu Wei.
\newblock Beit: {BERT} pre-training of image transformers.
\newblock In \emph{ICLR}, 2022.

\bibitem[{Bi} et~al.(2020){Bi}, {Chadha}, {Abbas}, {Bourtsoulatze}, and {Andreopoulos}]{Graph-based}
Y. {Bi}, A. {Chadha}, A. {Abbas}, E. {Bourtsoulatze}, and Y. {Andreopoulos}.
\newblock Graph-based spatio-temporal feature learning for neuromorphic vision sensing.
\newblock \emph{IEEE TIP}, pages 1--1, 2020.

\bibitem[Binas et~al.(2017)Binas, Neil, Liu, and Delbr{\"{u}}ck]{ddd17}
Jonathan Binas, Daniel Neil, Shih{-}Chii Liu, and Tobi Delbr{\"{u}}ck.
\newblock {DDD17:} end-to-end {DAVIS} driving dataset.
\newblock \emph{CoRR}, abs/1711.01458, 2017.

\bibitem[Brandli et~al.(2014)Brandli, Berner, Yang, Liu, and Delbruck]{brandli2014240}
Christian Brandli, Raphael Berner, Minhao Yang, Shih-Chii Liu, and Tobi Delbruck.
\newblock A 240$\times$ 180 130 db 3 $\mu$s latency global shutter spatiotemporal vision sensor.
\newblock \emph{JSSC}, 49\penalty0 (10):\penalty0 2333--2341, 2014.

\bibitem[Cannici et~al.(2019)Cannici, Ciccone, Romanoni, and Matteucci]{cannici2019asynchronous}
Marco Cannici, Marco Ciccone, Andrea Romanoni, and Matteo Matteucci.
\newblock Asynchronous convolutional networks for object detection in neuromorphic cameras.
\newblock In \emph{CVPRW}, 2019.

\bibitem[Cannici et~al.(2020)Cannici, Ciccone, Romanoni, and Matteucci]{Cannici_2020_ECCV}
Marco Cannici, Marco Ciccone, Andrea Romanoni, and Matteo Matteucci.
\newblock A differentiable recurrent surface for asynchronous event-based data.
\newblock In \emph{ECCV}, 2020.

\bibitem[Cheng et~al.(2020)Cheng, Luo, Yang, Yu, and Li]{CIFAR10DVS}
Wensheng Cheng, Hao Luo, Wen Yang, Lei Yu, and Wei Li.
\newblock Structure-aware network for lane marker extraction with dynamic vision sensor.
\newblock \emph{CoRR}, abs/2008.06204, 2020.

\bibitem[Deng et~al.(2021)Deng, Chen, and Li]{mvfnet}
Yongjian Deng, Hao Chen, and Youfu Li.
\newblock Mvf-net: A multi-view fusion network for event-based object classification.
\newblock \emph{IEEE TCSVT}, pages 1--1, 2021.

\bibitem[Deng et~al.(2022)Deng, Chen, Liu, and Li]{deng2022voxel}
Yongjian Deng, Hao Chen, Hai Liu, and Youfu Li.
\newblock A voxel graph cnn for object classification with event cameras.
\newblock In \emph{CVPR}, pages 1172--1181, 2022.

\bibitem[Deng et~al.(2023)Deng, Chen, Xie, Liu, and Li]{deng2023dynamic}
Yongjian Deng, Hao Chen, Bochen Xie, Hai Liu, and Youfu Li.
\newblock A dynamic graph cnn with cross-representation distillation for event-based recognition.
\newblock \emph{arXiv preprint arXiv:2302.04177}, 2023.

\bibitem[Dosovitskiy et~al.(2021)Dosovitskiy, Beyer, Kolesnikov, Weissenborn, Zhai, Unterthiner, Dehghani, Minderer, Heigold, Gelly, Uszkoreit, and Houlsby]{vit}
Alexey Dosovitskiy, Lucas Beyer, Alexander Kolesnikov, Dirk Weissenborn, Xiaohua Zhai, Thomas Unterthiner, Mostafa Dehghani, Matthias Minderer, Georg Heigold, Sylvain Gelly, Jakob Uszkoreit, and Neil Houlsby.
\newblock An image is worth 16x16 words: Transformers for image recognition at scale.
\newblock In \emph{ICLR}, 2021.

\bibitem[Gehrig et~al.(2019)Gehrig, Loquercio, Derpanis, and Scaramuzza]{gehrig2019end}
Daniel Gehrig, Antonio Loquercio, Konstantinos~G Derpanis, and Davide Scaramuzza.
\newblock End-to-end learning of representations for asynchronous event-based data.
\newblock In \emph{ICCV}, pages 5633--5643, 2019.

\bibitem[Gehrig et~al.(2021)Gehrig, Aarents, Gehrig, and Scaramuzza]{dsec}
Mathias Gehrig, Willem Aarents, Daniel Gehrig, and Davide Scaramuzza.
\newblock {DSEC:} {A} stereo event camera dataset for driving scenarios.
\newblock \emph{RAL}, 6\penalty0 (3):\penalty0 4947--4954, 2021.

\bibitem[Hagenaars et~al.(2021)Hagenaars, Paredes-Valles, and de~Croon]{hagenaars2021self}
Jesse Hagenaars, Federico Paredes-Valles, and Guido de Croon.
\newblock Self-supervised learning of event-based optical flow with spiking neural networks.
\newblock In \emph{NeurIPS}, pages 7167--7179. Curran Associates, Inc., 2021.

\bibitem[Han et~al.(2020)Han, Zhou, Duan, Tang, Xu, Xu, Huang, and Shi]{han2020neuromorphic}
Jin Han, Chu Zhou, Peiqi Duan, Yehui Tang, Chang Xu, Chao Xu, Tiejun Huang, and Boxin Shi.
\newblock Neuromorphic camera guided high dynamic range imaging.
\newblock In \emph{CVPR}, pages 1730--1739, 2020.

\bibitem[He et~al.(2022)He, Chen, Xie, Li, Doll{\'a}r, and Girshick]{he2022masked}
Kaiming He, Xinlei Chen, Saining Xie, Yanghao Li, Piotr Doll{\'a}r, and Ross Girshick.
\newblock Masked autoencoders are scalable vision learners.
\newblock In \emph{CVPR}, pages 16000--16009, 2022.

\bibitem[Huang et~al.(2023)Huang, Sun, Zhao, Li, and Su]{huang2023eventpoint}
Ze Huang, Li Sun, Cheng Zhao, Song Li, and Songzhi Su.
\newblock Eventpoint: Self-supervised interest point detection and description for event-based camera.
\newblock In \emph{WACV}, pages 5396--5405, 2023.

\bibitem[Kim et~al.(2021)Kim, Bae, Park, Zhang, and Kim]{nimagnet}
Junho Kim, Jaehyeok Bae, Gangin Park, Dongsu Zhang, and Young~Min Kim.
\newblock N-imagenet: Towards robust, fine-grained object recognition with event cameras.
\newblock In \emph{ICCV}, pages 2146--2156, 2021.

\bibitem[Klenk et~al.(2022)Klenk, Bonello, Koestler, and Cremers]{mem}
Simon Klenk, David Bonello, Lukas Koestler, and Daniel Cremers.
\newblock Masked event modeling: Self-supervised pretraining for event cameras.
\newblock \emph{arXiv preprint arXiv:2212.10368}, 2022.

\bibitem[Li et~al.(2021)Li, Zhou, Yang, Zhang, Cui, Bao, and Zhang]{Li21iccv}
Yijin Li, Han Zhou, Bangbang Yang, Ye Zhang, Zhaopeng Cui, Hujun Bao, and Guofeng Zhang.
\newblock Graph-based asynchronous event processing for rapid object recognition.
\newblock In \emph{ICCV}, pages 914--923, 2021.

\bibitem[{Lichtsteiner} et~al.(2008){Lichtsteiner}, {Posch}, and {Delbruck}]{lichtsteinerposch}
P. {Lichtsteiner}, C. {Posch}, and T. {Delbruck}.
\newblock A 128$\times$ 128 120 db 15 $\mu$s latency asynchronous temporal contrast vision sensor.
\newblock \emph{JSSC}, 43\penalty0 (2):\penalty0 566--576, 2008.

\bibitem[Loshchilov and Hutter(2016)]{Loshchilov_Hutter_2016}
Ilya Loshchilov and Frank Hutter.
\newblock Sgdr: Stochastic gradient descent with warm restarts.
\newblock \emph{arXiv preprint arXiv:1608.03983}, 2016.

\bibitem[Messikommer et~al.(2020)Messikommer, Gehrig, Loquercio, and Scaramuzza]{messikommer2020event}
Nico Messikommer, Daniel Gehrig, Antonio Loquercio, and Davide Scaramuzza.
\newblock Event-based asynchronous sparse convolutional networks.
\newblock In \emph{ECCV}, pages 415--431. Springer, 2020.

\bibitem[Min et~al.(2022)Min, Zhao, Xiao, Nie, and Dai]{min2022voxel}
Chen Min, Dawei Zhao, Liang Xiao, Yiming Nie, and Bin Dai.
\newblock Voxel-mae: Masked autoencoders for pre-training large-scale point clouds.
\newblock \emph{arXiv preprint arXiv:2206.09900}, 2022.

\bibitem[Orchard et~al.(2015)Orchard, Jayawant, Cohen, and Thakor]{ncaltech}
Garrick Orchard, Ajinkya Jayawant, Gregory Cohen, and Nitish~V. Thakor.
\newblock Converting static image datasets to spiking neuromorphic datasets using saccades.
\newblock \emph{CoRR}, abs/1507.07629, 2015.

\bibitem[Pang et~al.(2022)Pang, Wang, Tay, Liu, Tian, and Yuan]{pang2022masked}
Yatian Pang, Wenxiao Wang, Francis~EH Tay, Wei Liu, Yonghong Tian, and Li Yuan.
\newblock Masked autoencoders for point cloud self-supervised learning.
\newblock In \emph{ECCV}, pages 604--621. Springer, 2022.

\bibitem[Paredes-Vall{\'e}s and de~Croon(2021)]{paredes2021back}
Federico Paredes-Vall{\'e}s and Guido~CHE de Croon.
\newblock Back to event basics: Self-supervised learning of image reconstruction for event cameras via photometric constancy.
\newblock In \emph{CVPR}, pages 3446--3455, 2021.

\bibitem[Peng et~al.(2023)Peng, Zhang, Xiong, Sun, and Wu]{peng2023get}
Yansong Peng, Yueyi Zhang, Zhiwei Xiong, Xiaoyan Sun, and Feng Wu.
\newblock Get: Group event transformer for event-based vision.
\newblock In \emph{ICCV}, pages 6038--6048, 2023.

\bibitem[Qi et~al.(2017)Qi, Yi, Su, and Guibas]{qi2017pointnet++}
Charles~Ruizhongtai Qi, Li Yi, Hao Su, and Leonidas~J Guibas.
\newblock Pointnet++: Deep hierarchical feature learning on point sets in a metric space.
\newblock In \emph{NeurIPS}, 2017.

\bibitem[Rebecq et~al.(2019)Rebecq, Ranftl, Koltun, and Scaramuzza]{rebecq2019high}
Henri Rebecq, Ren{\'e} Ranftl, Vladlen Koltun, and Davide Scaramuzza.
\newblock High speed and high dynamic range video with an event camera.
\newblock \emph{IEEE TPAMI}, 43\penalty0 (6):\penalty0 1964--1980, 2019.

\bibitem[Schaefer et~al.(2022)Schaefer, Gehrig, and Scaramuzza]{Schaefer22cvpr}
Simon Schaefer, Daniel Gehrig, and Davide Scaramuzza.
\newblock Aegnn: Asynchronous event-based graph neural networks.
\newblock In \emph{CVPR}, 2022.

\bibitem[Sun et~al.(2022)Sun, Messikommer, Gehrig, and Scaramuzza]{ess}
Zhaoning Sun, Nico Messikommer, Daniel Gehrig, and Davide Scaramuzza.
\newblock {ESS:} learning event-based semantic segmentation from still images.
\newblock In \emph{ECCV}, pages 341--357. Springer, 2022.

\bibitem[Tong et~al.(2022)Tong, Song, Wang, and Wang]{tong2022videomae}
Zhan Tong, Yibing Song, Jue Wang, and Limin Wang.
\newblock Videomae: Masked autoencoders are data-efficient learners for self-supervised video pre-training.
\newblock In \emph{NeurIPS}, pages 10078--10093, 2022.

\bibitem[Wang et~al.(2020)Wang, He, Yu, Xia, and Yang]{wang2020event}
Bishan Wang, Jingwei He, Lei Yu, Gui-Song Xia, and Wen Yang.
\newblock Event enhanced high-quality image recovery.
\newblock In \emph{ECCV}, pages 155--171. Springer, 2020.

\bibitem[Wang et~al.(2023{\natexlab{a}})Wang, Weng, Zhang, and Xiong]{wang2023unsupervised}
Jin Wang, Wenming Weng, Yueyi Zhang, and Zhiwei Xiong.
\newblock Unsupervised video deraining with an event camera.
\newblock In \emph{ICCV}, pages 10831--10840, 2023{\natexlab{a}}.

\bibitem[Wang et~al.(2023{\natexlab{b}})Wang, Huang, Zhao, Tong, He, Wang, Wang, and Qiao]{wang2023videomae}
Limin Wang, Bingkun Huang, Zhiyu Zhao, Zhan Tong, Yinan He, Yi Wang, Yali Wang, and Yu Qiao.
\newblock Videomae v2: Scaling video masked autoencoders with dual masking.
\newblock In \emph{CVPR}, pages 14549--14560, 2023{\natexlab{b}}.

\bibitem[Wei et~al.(2022)Wei, Fan, Xie, Wu, Yuille, and Feichtenhofer]{wei2022masked}
Chen Wei, Haoqi Fan, Saining Xie, Chao-Yuan Wu, Alan Yuille, and Christoph Feichtenhofer.
\newblock Masked feature prediction for self-supervised visual pre-training.
\newblock In \emph{CVPR}, pages 14668--14678, 2022.

\bibitem[Wu et~al.(2022)Wu, Zhang, Lin, Li, Wang, and Tang]{LIAF-Net}
Zhenzhi Wu, Hehui Zhang, Yihan Lin, Guoqi Li, Meng Wang, and Ye Tang.
\newblock Liaf-net: Leaky integrate and analog fire network for lightweight and efficient spatiotemporal information processing.
\newblock \emph{TNNLS}, 33\penalty0 (11):\penalty0 6249--6262, 2022.

\bibitem[Xie et~al.(2022{\natexlab{a}})Xie, Deng, Shao, Liu, and Li]{Xie_2022_RAL}
Bochen Xie, Yongjian Deng, Zhanpeng Shao, Hai Liu, and Youfu Li.
\newblock Vmv-gcn: Volumetric multi-view based graph cnn for event stream classification.
\newblock \emph{RAL}, 7\penalty0 (2):\penalty0 1976--1983, 2022{\natexlab{a}}.

\bibitem[Xie et~al.(2022{\natexlab{b}})Xie, Zhang, Cao, Lin, Bao, Yao, Dai, and Hu]{xie2022simmim}
Zhenda Xie, Zheng Zhang, Yue Cao, Yutong Lin, Jianmin Bao, Zhuliang Yao, Qi Dai, and Han Hu.
\newblock Simmim: A simple framework for masked image modeling.
\newblock In \emph{CVPR}, pages 9653--9663, 2022{\natexlab{b}}.

\bibitem[Yang et~al.(2023{\natexlab{a}})Yang, He, Liu, Chen, Wu, Lin, He, and Ouyang]{yang2023gd}
Honghui Yang, Tong He, Jiaheng Liu, Hua Chen, Boxi Wu, Binbin Lin, Xiaofei He, and Wanli Ouyang.
\newblock Gd-mae: generative decoder for mae pre-training on lidar point clouds.
\newblock In \emph{CVPR}, pages 9403--9414, 2023{\natexlab{a}}.

\bibitem[Yang et~al.(2023{\natexlab{b}})Yang, Pan, and Liu]{yang2023event}
Yan Yang, Liyuan Pan, and Liu Liu.
\newblock Event camera data pre-training.
\newblock \emph{ICCV}, 2023{\natexlab{b}}.

\bibitem[Yao et~al.(2021)Yao, Gao, Zhao, Wang, Lin, Yang, and Li]{yao2021temporal}
Man Yao, Huanhuan Gao, Guangshe Zhao, Dingheng Wang, Yihan Lin, Zhaoxu Yang, and Guoqi Li.
\newblock Temporal-wise attention spiking neural networks for event streams classification.
\newblock In \emph{ICCV}, pages 10221--10230, 2021.

\bibitem[Zhang et~al.(2022)Zhang, Guo, Gao, Fang, Zhao, Wang, Qiao, and Li]{zhang2022point}
Renrui Zhang, Ziyu Guo, Peng Gao, Rongyao Fang, Bin Zhao, Dong Wang, Yu Qiao, and Hongsheng Li.
\newblock Point-m2ae: Multi-scale masked autoencoders for hierarchical point cloud pre-training.
\newblock In \emph{NeurIPS}, pages 27061--27074, 2022.

\end{thebibliography}
\end{document}